\documentclass{article} 
\pdfoutput=1
\usepackage{nips13submit_e,times}
\usepackage{url}
\usepackage[numbers]{natbib}

\usepackage{times}
\usepackage{latexsym}

\usepackage{amsmath}
\usepackage{color}
\usepackage{amsfonts} 
\usepackage{graphicx}
\usepackage{xspace}
\usepackage{subcaption} 

\usepackage[export]{adjustbox} 

\usepackage{tikz}
\usepackage{pgfplots}

\newcommand{\biCVM}{\textsc{biCVM}\xspace}
\newcommand{\biCVMplus}{\textsc{biCVM+}\xspace}

\newcommand{\CVM}{\textsc{CVM}\xspace}

\newcommand\kmh[1]{}
\newcommand\pb[1]{}

\mathchardef\mhyphen="2D

\title{Multilingual Distributed Representations without Word Alignment}

\author{Karl Moritz Hermann and Phil Blunsom\\
 Department of Computer Science\\
 University of Oxford\\
 Oxford, OX1 3QD, UK\\
 {\tt \{karl.moritz.hermann,phil.blunsom\}@cs.ox.ac.uk}}

\date{}
\nipsfinalcopy

\begin{document}
\maketitle
\begin{abstract}
Distributed representations of meaning are a natural way to encode covariance
relationships between words and phrases in NLP.  By overcoming data sparsity
problems, as well as providing information about semantic relatedness which is
not available in discrete representations, distributed representations have
proven useful in many NLP tasks. Recent work has shown how compositional
semantic representations can successfully be applied to a number of monolingual
applications such as sentiment analysis.  At the same time, there has been some
initial success in work on learning shared word-level representations across
languages.  We combine these two approaches by proposing a method for learning
distributed representations in a multilingual setup.  Our model learns to assign
similar embeddings to aligned sentences and dissimilar ones to sentence which
are not aligned while not requiring word alignments.  We show that our
representations are semantically informative and apply them to a cross-lingual
document classification task where we outperform the previous state of the art.
Further, by employing parallel corpora of multiple language pairs we find that
our model learns representations that capture semantic relationships across
languages for which no parallel data was used.
\end{abstract}

\section{Introduction}

Distributed representations of words are increasingly being used to achieve high levels of generalisation within language modelling tasks.
Successful applications of this approach include word-sense disambiguation, word similarity and synonym detection (e.g. \cite{Erk:2008,Turney:2010}).
Subsequent work has also attempted to learn distributed semantics of larger structures, allowing us to apply distributed representation to tasks such as sentiment analysis or paraphrase detection (i.a. \cite{Baroni:2010,Clark:2007a,Grefenstette:2011,Hermann:2013:ACL,Mitchell:2008,Socher:2012}).
At the same time a second strand of work has focused on transferring linguistic knowledge across languages, and particularly from English into low-resource languages, by means of distributed representations at the word level \cite{Haghighi:2008,Klementiev:2012}.

Currently, work on compositional semantic representations focuses on monolingual data while the cross-lingual work focuses on word level representations only.
However, it appears logical that these two strands of work should be combined as there exists a plethora of parallel corpora with aligned data at the sentence level or beyond which could be exploited in such work.
Further, sentence aligned data provides a plausible concept of semantic similarity, which can be harder to define at the word level.
Consider the case of alignment between a German compound noun (e.g. ``Schwerlastverkehr'') and its English equivalent (``heavy goods vehicle traffic''). 
Semantic alignment at the phrase level here appears far more plausible than aligning individual tokens for semantic transfer.

Using this rationale, and building on both work related to learning cross-lingual embeddings as well as to compositional semantic representations, we introduce a model that learns cross-lingual embeddings at the sentence level.
In the following section we will briefly discuss prior work in these two fields before going on to describe the bilingual training signal that we developed for learning multilingual compositional embeddings.
Subsequently, we will describe our model in greater detail as well as its training procedure and experimental setup.
Finally, we perform a number of evaluations and demonstrate that our training signal allows a very simple compositional vector model to outperform the state of the art on a task designed to evaluate its ability to transfer semantic information across languages.
Unlike other work in this area, our model does not require word aligned data.
In fact, while we evaluate our model on sentence aligned data in this paper, there is no theoretical requirement for this and technically our algorithm could also be applied to document-level parallel data or even comparable data only.
\section{Models of Compositional Distributed Semantics}

In the case of representing individual words as vectors, the distributional
account of semantics provides a plausible explanation of what is encoded in a
word vector.  This follows the idea that the meaning of a word can be determined
by ``the company it keeps'' \cite{Firth:1957}, that is by the context it appears
in.  Such context can easily be encoded in vectors using collocational methods,
and is also underlying other methods of learning word embeddings
\cite{Collobert:2008,Mikolov:2010}.

For a number of important problems, semantic representations of individual words
do not suffice, but instead a semantic representation of a larger
structure---e.g. a phrase or a sentence---is required.  This was highlighted in
\cite{Erk:2008}, who proposed a mechanism for modifying a word's representation
based on its individual context.  The distributional account of semantics can,
due to sparsity, not be applied to such larger linguistic units.  A notable
exception perhaps is \citet{Baroni:2010}, who learned distributional
representations for adjective noun pairs using a collocational approach on a
corpus of unprecedented size.  The bigram representations learned from that
corpus were subsequently used to learn lexicalised composition functions for the
constituent words.

Most alternative attempts to extract such higher-level semantic representations
have focused on learning composition functions that represent the semantics of a
larger structure as a function of the representations of its parts.
\cite{Mitchell:2008} provides an evaluation of a number of simple composition
functions applied to bigrams.  Applied recursively, such approaches can then
easily be reconciled with the co-occurrence based word level representations.
There are a number of proposals motivating such recursive or deep composition
models.  Notably, \cite{Clark:2007a} propose a tensor-based model for semantic
composition and, similarly, \cite{Coecke:2010} develop a framework for semantic
composition by combining distributional theory with pregroup grammars.  The
latter framework was empirically evaluated and supported by the results in
\cite{Grefenstette:2011}.  More recently, various forms of recursive neural
networks have successfully been used for semantic composition and related tasks
such as sentiment analysis.  Such models include recursive autoencoders
\cite{Socher:2011}, matrix-vector recursive neural networks \cite{Socher:2012},
untied recursive neural networks \cite{Hermann:2013:ACL} or convolutional
networks \cite{Kalchbrenner:2013}.

\subsection{Multilingual Embeddings}

Much research has been devoted to the task of inducing distributed semantic
representations for single languages.  In particular English, with its large
number of annotated resources, has enjoyed most attention.  Recently, progress
has been made at representation learning for languages with fewer available
resources.  \citet{Klementiev:2012} described a form of multitask learning on
word-aligned parallel data to transfer embeddings from one language to another.
Earlier work, \citet{Haghighi:2008}, proposed a method for inducing
cross-lingual lexica using monolingual feature representations and a small
initial lexicon to bootstrap with.  This approach has recently been extended by
\cite{Mikolov:2013,Mikolov:2013a}, who developed a method for learning
transformation matrices to convert semantic vectors of one language into those
of another.  Is was demonstrated that this approach can be applied to improve
tasks related to machine translation.  Their CBOW model is also worth noting for
its similarities to the composition function used in this paper.  Using a
slightly different approach, \cite{Zou:2013}, also learned bilingual embeddings
for machine translation.  It is important to note that, unlike our proposed
system, all of these methods require word aligned parallel data for training.

Two recent workshop papers deserve mention in this respect. Both
\citet{Lauly:2013} and \citet{Chandar:2013} propose methods for learning
word embeddings by exploiting bilingual data, not unlike the method proposed in
this paper. Instead of the noise-contrastive method developed in this paper,
both groups of authors make use of autoencoders to encode monolingual
representations and to support the bilingual transfer.

So far almost all of this work has been focused on learning multilingual
representations at the word level. As distributed representations of larger
expressions have been shown to be highly useful for a number of tasks, it seems
to be a natural next step to also attempt to induce these using cross-lingual
data. This paper provides a first step in that direction.

\section{Model Description}

Language acquisition in humans is widely seen as grounded in sensory-motor experience \cite{Roy:2003,Bloom:2001}.
Based on this idea, there have been some attempts at using multi-modal data for learning better vector representations of words (e.g. \cite{Srivastava:2012}).
Such methods, however, are not easily scalable across languages or to large amounts of data for which no secondary or tertiary representation might exist.

We abstract the underlying principle one step further and attempt to learn semantics from multilingual data.
The idea is that, given enough parallel data, a shared representation would be forced to capture the common elements between sentences from different languages.
What two parallel sentences have in common, of course, is the semantics of those two sentences.
Using this data, we propose a novel method for learning vector representations at the word level and beyond.

\subsection{Bilingual Signal}

Exploiting the semantic similarity of parallel sentences across languages, we
can define a simple bilingual (and trivially multilingual) error function as
follows: Given a compositional sentence model (\CVM) $\mathcal{M}_A$, which maps
a sentence to a vector, we can train a second \CVM $\mathcal{M}_B$ using a
corpus $\mathcal{C}_{A,B}$ of parallel data from the language pair $A,B$.  For
each pair of parallel sentences $(a,b) \in \mathcal{C}_{A,B}$, we attempt to
minimize
\begin{align}
E_{dist}(a,b) = \left\| a_{root} - b_{root} \right\|^2\label{eqn:bi-error}
\end{align}
where $a_{root}$ is the vector representing sentence $a$ and $b_{root}$ the vector representing sentence $b$.

\subsection{The \biCVM Model}

\begin{figure}[t]
\begin{center}
\includegraphics[height=200pt]{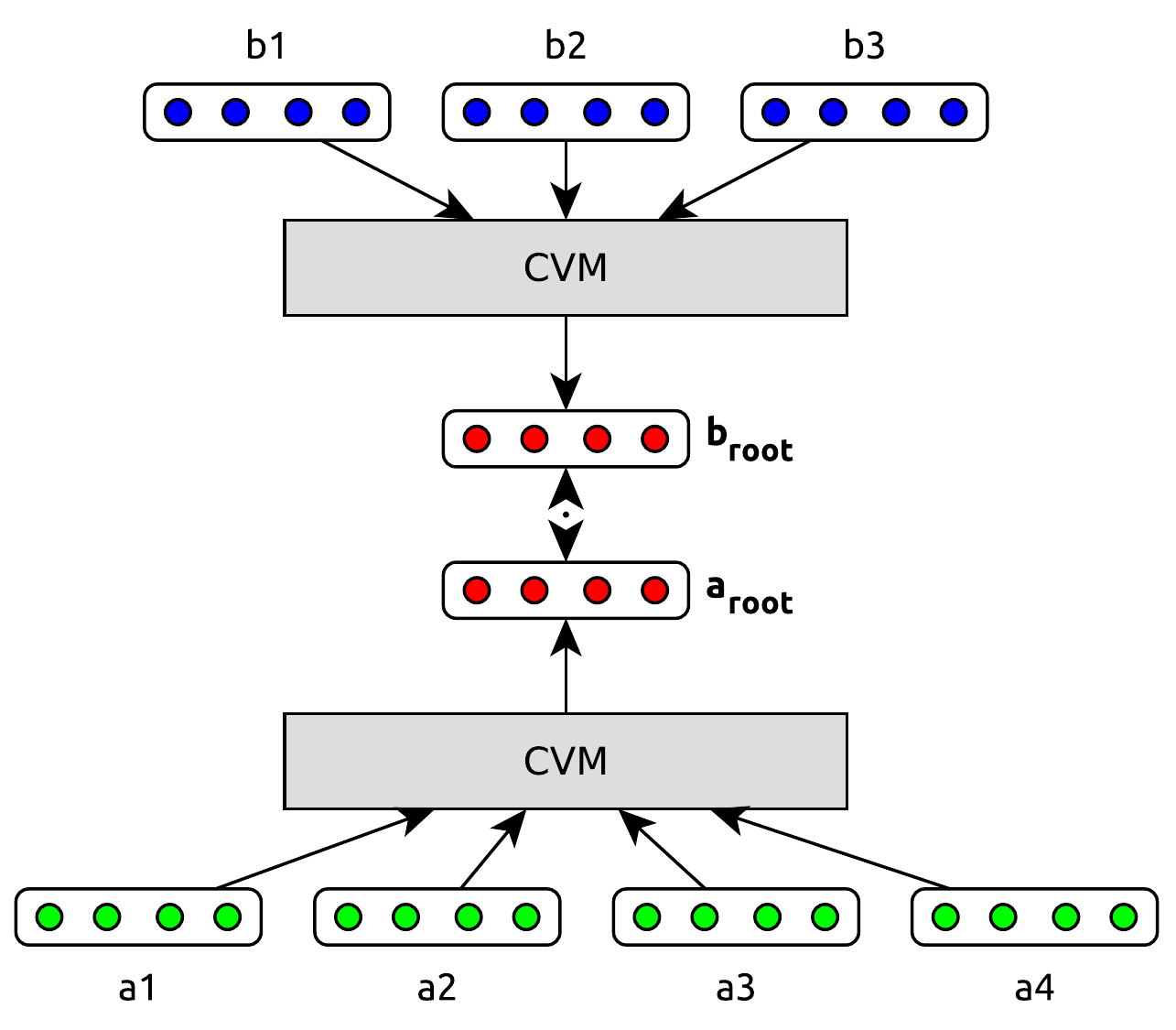}
\caption{Description of a bilingual model with parallel input sentences $a$ and $b$. The objective function of this model is to minimize the distance between the sentence level encoding of the bitext. Principally any composition function can be used to generate the compositional sentence level representations. The composition function is represented by the \textit{CVM} boxes in the diagram above.
}\label{fig:bilingual}
\end{center}
\end{figure}

A \CVM learns semantic representations of larger syntactic units given the semantic representations of their constituents.
We assume individual words to be represented by vectors ($x\in\mathbb{R}^d$).

Previous methods employ binary parse trees on the data (e.g. \cite{Hermann:2013:ACL,Socher:2012}) and use weighted or multiplicative composition functions.
Under such a setup, where each node in the tree is terminal or has two children ($p \rightarrow c_0,c_1$), a binary composition function could take the following form:
\begin{align}
&p = g\left(W^e [c_0;c_1] + b^e \right) \label{eq:encoding}
\end{align}
where $[c_0;c_1]$ is the concatenation of the two child vectors, $W^e \in
\mathbb{R}^{d\times 2d}$ and $b^e \in \mathbb{R}^d$ the encoding matrix and bias, respectively, and $g$ an element-wise activation function such as the hyperbolic tangent.
For the purposes of evaluation the bilingual signal proposed above, we simplify this composition function by setting all weight matrices to the identity and all biases to zero.
Thereby the \CVM reduces to a simple additive composition function:
\begin{align}
a_{root} = \sum_{i=0}^{|a|} a_i
\end{align}
Of course, this is a very simplified \CVM, as such a bag-of-words approach no
longer accounts for word ordering and other effects which a more complex \CVM
might capture. However, for the purposes of this evaluation (and with the
experimental evaluation in mind), such a simplistic composition function should
be sufficient to evaluate the novel objective function proposed here.

Using this additive \CVM we want to optimize the bilingual error signal defined above (Eq. \ref{eqn:bi-error}).
For the moment, assume that $\mathcal{M}_A$ is a perfectly trained \CVM such that $a_{root}$ represents the semantics of the sentence $a$.
Further, due to the use of parallel data, we know that $a$ and $b$ are semantically equivalent.
Hence we transfer the semantic knowledge contained in $\mathcal{M}_A$ onto $\mathcal{M}_B$, by learning $\theta_{\mathcal{M}_B}$ to minimize:
\begin{align}
E_{bi}(\mathcal{C}_{A,B}) = \sum_{(a,b) \in \mathcal{C}_{A,B}} E_{dist}(a,b)
\end{align}

Of course, this objective function assumes a fully trained model which we do not have at this stage.
While this can be a useful objective for transferring linguistic knowledge into low-resource languages \cite{Klementiev:2012}, this precondition is not helpful when there is no model to learn from in first place.
We resolve this issue by jointly training both models $\mathcal{M}_A$ and $\mathcal{M}_B$.

Applying $E_{bi}$ to parallel data ensures that both models learn a shared representation at the sentence level.
As the parallel input sentences share the same meaning, it is reasonable to assume that minimizing $E_{bi}$ will force the model to learn their semantic representation.
Let $\theta_{bi} = \theta_{\mathcal{M}_A} \cup \theta_{\mathcal{M}_B}$.
The joint objective function $J(\theta_{bi})$ thus becomes:
\begin{align}
J(\theta_{bi}) = E_{bi}(\mathcal{C}_{A,B}) + \frac{\lambda}{2}\|\theta_{bi}\|^2
\end{align}
where $\lambda\|\theta_{bi}\|_1$ is the $L_2$ regularization term.

It is apparent that this joint objective $J(\theta_{bi})$ is degenerate.
The models could learn to reduce all embeddings and composition weights to zero and thereby minimize the objective function.
We address this issue by employing a form of contrastive estimation penalizing small distances between non-parallel sentence pairs.
For every pair of parallel sentences $(a,b)$ we sample a number of additional sentences $n \in \mathcal{C}_B$, which---with high probability---are not exact translations of $a$.
This is comparable to the second term of the loss function of a large margin nearest neighbour classifier (see Eq. 12 in \cite{Weinberger:2009}):
\begin{align}
E_{noise}(a,b,n) = \left[1 + E_{dist}(a,b) - E_{dist}(a,n)\right]_{+}
\end{align}
where $[x]_{+} = max(x,0)$ denotes the standard hinge loss.
Thus, the final objective function to minimize for the \biCVM model is:
\begin{align}
  J(\theta_{bi})=\sum_{(a,b) \in \mathcal{C}_{A,B}} \left( \sum_{i=1}^{k}
    E_{noise}(a,b,n_i)\right) + \frac{\lambda}{2}\|\theta_{bi}\|^2
\end{align}

\subsection{Model Learning}

Given the objective function as defined above, model learning can employ the
same techniques as any monolingual \CVM.  In particular, as the objective
function is differentiable, we can use standard gradient descent techniques such
as stochastic gradient descent, L-BFGS or the adaptive gradient algorithm
AdaGrad \cite{Duchi:2011}. Within each monolingual \CVM, we use backpropagation
through structure after applying the joint error to each sentence level node.

\section{Experiments}

\subsection{Data and Parameters}

All model weights were randomly initialised using a Gaussian distribution.
There are a number of parameters that can influence model training. We selected
the following values for simplicity and comparability with prior work. In future
work we will investigate the effect of these parameters in greater detail. L2
regularization (1), step-size (0.1), number of noise elements (50), margin size
(50), embedding dimensionality ($d{=}40$).  The noise elements samples were
randomly drawn from the corpus at training time, individually for each training
sample and epoch.  We use the Europarl corpus
(v7)\footnote{\url{http://www.statmt.org/europarl/}} for training the bilingual
model.  The corpus was pre-processed using the set of tools provided by
cdec\footnote{\url{https://github.com/redpony/cdec}} \cite{Dyer:2010} for
tokenizing and lowercasing the data. Further, all empty sentences as well as
their translations were removed from the corpus.

We present results from two experiments.  The \biCVM model was trained on 500k
sentence pairs of the English-German parallel section of the Europarl corpus.
The \biCVMplus model used this dataset in combination with another 500k parallel
sentences from the English-French section of the corpus, resulting in 1 million
English sentences, each paired up with either a German or a French sentence.
Each language's vocabulary used distinct encodings to avoid potential overlap.

The motivation behind \biCVMplus is to investigate whether we can learn better
embeddings by introducing additional data in a different language.  This is
similar to prior work in machine translation where English was used as a pivot
for translation between low-resource languages \cite{Cohn:2007}.

We use the adaptive gradient method, AdaGrad \cite{Duchi:2011}, for updating the
weights of our models, and terminate training after 50 iterations. Earlier
experiments indicated that the \biCVM model converges faster than the \biCVMplus
model, but we report results on the same number of iterations for better
comparability\footnote{These numbers were updated following comments in the ICLR
  open review process. Results for other dimensionalities and our source code
  for our model are available at \url{http://www.karlmoritz.com}.}.

\subsection{Cross-Lingual Document Classification}

We evaluate our model using the cross-lingual document classification (CLDC)
task of \citet{Klementiev:2012}.  This task involves learning language
independent embeddings which are then used for document classification across
the English-German language pair.  For this, CLDC employs a particular kind of
supervision, namely using supervised training data in one language and
evaluating without supervision in another.  Thus, CLDC is a good task for
establishing whether our learned representations are semantically useful across
multiple languages.

We follow the experimental setup described in \cite{Klementiev:2012}, with the
exception that we learn our embeddings using solely the Europarl data and only
use the Reuters RCV1/RCV2 corpora during the classifier training and testing
stages.  Each document in the classification task is represented by the average
of the $d$-dimensional representations of all its sentences.  We train the
multiclass classifier using the same settings and implementation of the averaged
perceptron classifier \cite{Collins:2002} as used in \cite{Klementiev:2012}.

\begin{table}[t]
\centering
\begin{tabular}{lcc}
Model & en $\rightarrow$ de & de $\rightarrow$ en \\ \hline
Majority Class & 46.8 & 46.8 \\
Glossed & 65.1 & 68.6 \\
MT & 68.1 & 67.4 \\
I-Matrix & 77.6 & 71.1 \\ \hline
\biCVM & 83.7 & 71.4 \\
\biCVMplus & \textbf{86.2} & \textbf{76.9} \\
\end{tabular}
\caption{Classification accuracy for training on English and German with 1000
  labeled examples. Cross-lingual compositional representations (\biCVM and
  \biCVMplus), cross-lingual representations using learned embeddings and an
  interaction matrix (I-Matrix) \cite{Klementiev:2012} translated (MT) and
  glossed (Glossed) words, and the majority class baseline. The MT and Glossed
  results are also taken from \citet{Klementiev:2012}.}
\label{tab:results1k}
\end{table}

\pgfplotsset{every axis plot/.append style={line width=1pt}}

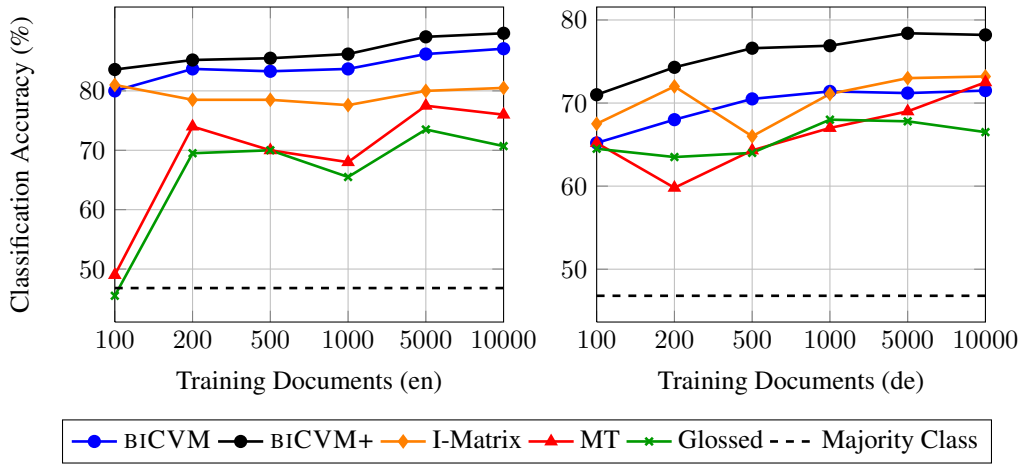
\begin{figure}[t]
\begin{tikzpicture}
	\begin{axis}[
		grid=major,
		ytick={40,50,...,80},
		xlabel=Training Documents (en),
    		xtick={1,2,3,4,5,6},
	    xticklabels={$100$,$200$,$500$,$1000$,$5000$,$10000$},
		ylabel=Classification Accuracy (\%),
		legend columns=-1,
		legend entries={\biCVM, \biCVMplus, I-Matrix, MT, Glossed, Majority Class},
		legend to name=sharedlegend,
		scale only axis,
		xmin=1,xmax=6,
		width=0.37\textwidth,
		height=0.3\textwidth,
		]
	\addplot[color=blue,mark=*] coordinates {
	(1, 80.0)
	(2, 83.7)
	(3, 83.3)
	(4, 83.7)
	(5, 86.2)
	(6, 87.1)
	};
	\addplot[color=black,mark=*] coordinates {
	(1, 83.6)
	(2, 85.2)
	(3, 85.5)
	(4, 86.2)
	(5, 89.1)
	(6, 89.7)
	};
	\addplot[color=orange,mark=diamond*] coordinates {
	(1, 81)
	(2, 78.5)
	(3, 78.5)
	(4, 77.6)
	(5, 80)
	(6, 80.5)
	};
	\addplot[color=red,mark=triangle*] coordinates {
	(1, 49)
	(2, 74)
	(3, 70)
	(4, 68)
	(5, 77.5)
	(6, 76)
	};
	\addplot[color=green!60!black,mark=x] coordinates {
	(1, 45.5)
	(2, 69.5)
	(3, 70)
	(4, 65.5)
	(5, 73.5)
	(6, 70.7)
	};
	\addplot[color=black,style=dashed] coordinates {
	(1, 46.8)
	(6, 46.8)
	};
	\end{axis}
\end{tikzpicture}
\begin{tikzpicture}
	\begin{axis}[
		grid=major,
		xlabel=Training Documents (de),
    		xtick={1,2,3,4,5,6},
	    xticklabels={$100$,$200$,$500$,$1000$,$5000$,$10000$},
		scale only axis,
		xmin=1,xmax=6,
		width=0.37\textwidth,
		height=0.3\textwidth,
		]
	\addplot[color=blue,mark=*] coordinates {
	(1, 65.2)
	(2, 68.0)
	(3, 70.5)
	(4, 71.4)
	(5, 71.2)
	(6, 71.5)
	};
	\addplot[color=black,mark=*] coordinates {
	(1, 71.0)
	(2, 74.3)
	(3, 76.6)
	(4, 76.9)
	(5, 78.4)
	(6, 78.2)
	};
	\addplot[color=orange,mark=diamond*] coordinates {
	(1, 67.5)
	(2, 72)
	(3, 66)
	(4, 71.1)
	(5, 73)
	(6, 73.2)
	};
	\addplot[color=red,mark=triangle*] coordinates {
	(1, 65.2)
	(2, 59.8)
	(3, 64.3)
	(4, 67)
	(5, 69)
	(6, 72.5)
	};
	\addplot[color=green!60!black,mark=x] coordinates {
	(1, 64.5)
	(2, 63.5)
	(3, 64)
	(4, 68)
	(5, 67.8)
	(6, 66.5)
	};
	\addplot[color=black,style=dashed] coordinates {
	(1, 46.8)
	(6, 46.8)
	};
	\end{axis}
\end{tikzpicture}

\begin{center}
\ref{sharedlegend}
\end{center}
\caption{Classification accuracy for a number of models (see Table
  \ref{tab:results1k} for model descriptions). The left chart shows results for
  these models when trained on English data and evaluated on German data, the
  right chart vice versa.}\label{fig:cldccharts}
\end{figure}

We ran the CLDC experiments both by training on English and testing on German
documents and vice versa.  Using the data splits provided by
\cite{Klementiev:2012}, we used varying training data sizes from 100 to 10,000
documents for training the multiclass classifier.  The results of this task
across training sizes are shown in Figure \ref{fig:cldccharts}.  Table
\ref{tab:results1k} shows the results for training on 1,000 documents.

Both models, \biCVM and \biCVMplus outperform all prior work on this task.
Further, the \biCVMplus model outperforms the \biCVM model, indicating the
usefulness of adding training data even from a separate language pair.

\subsection{Visualization}

\begin{figure}[t]
\includegraphics[scale=0.38, frame]{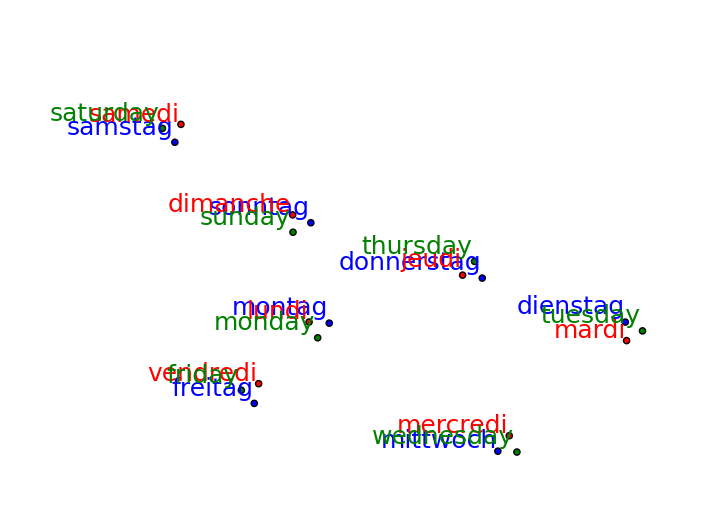}
\includegraphics[scale=0.38, frame]{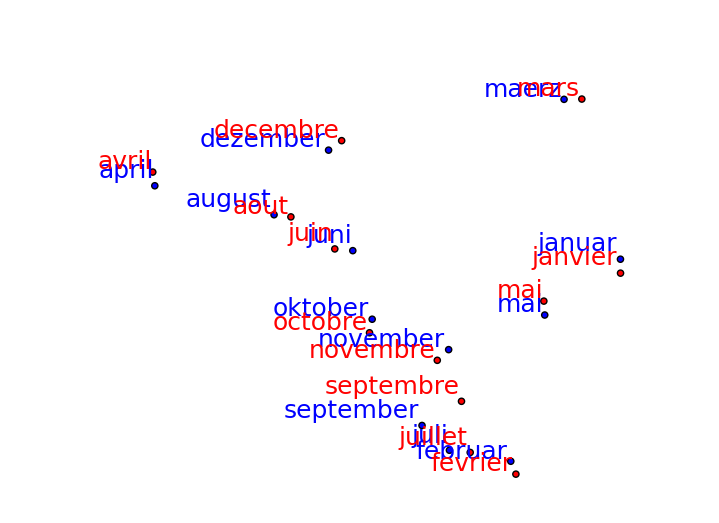}
\caption{The left scatter plot shows t-SNE projections for a weekdays in all
  three languages using the representations learned in the \biCVMplus model.
  Even though the model did not use any parallel French-German data during
  training, it still learns semantic similarity between these two languages
  using English as a pivot. To highlight this, the right plot shows another set
  of words (months of the year) using only the German and French
  words.}\label{fig:scatterdefr}
\end{figure}

While the CLDC experiment focused on establishing the semantic content of the
sentence level representations, we also want to briefly investigate the induced
word embeddings.  In particular the \biCVMplus model is interesting for that
purpose, as it allows us to evaluate our approach of using English as a pivot
language in a multilingual setup.

In Figure \ref{fig:scatterdefr} we show the t-SNE projections for a number of
English, French and German words.  Of particular interest should be the right
chart, which highlights bilingual embeddings between French and German words.
Even though the model did not use any parallel French-German data during
training, it still managed to learn semantic word-word similarity across these
two languages.

\section{Conclusions}

With this paper we have proposed a novel method for inducing cross-lingual
distributed representations for compositional semantics.  Using a very simple
method for semantic composition, we nevertheless managed to obtain state of the
art results on the CLDC task, specifically designed to evaluate semantic
transfer across languages.  After extending our approach to include multilingual
training data in the \biCVMplus model, we were able to demonstrate that adding
additional languages further improves the model.  Furthermore, using some
qualitative experiments and visualizations, we showed that our approach also
allows us to learn semantically related embeddings across languages without any
direct training data.

Our approach provides great flexibility in training data and requires little to
no annotation.  Having demonstrated the successful training of semantic
representations using sentence aligned data, a plausible next step is to attempt
training using document-aligned data or even corpora of comparable documents.
This may provide even greater possibilities for working with low-resource
languages.

In the same vein, the success of our pivoting experiments suggest further work.
Unlike other pivot approaches, it is easy to extend our model to have multiple
pivot languages.  Thus some pivots could preserve different aspects such as
case, gender etc., and overcome other issues related to having a single pivot
language.

As we have achieved the results in this paper with a relatively simple \CVM, it
would also be interesting to establish whether our objective function can be
used in combination with more complex compositional vector
models such as MV-RNN \cite{Socher:2012} or tensor-based approaches, to see
whether these can further improve results on both mono- and multilingual tasks
when used in conjunction with our cross-lingual objective function.  Related to
this, we will also apply our model to a wider variety of tasks including machine
translation and multilingual information extraction.

\section*{Acknowledgements}

The authors would like to thank Alexandre Klementiev and his co-authors for
making their datasets and averaged perceptron implementation available, as well
as answering a number of questions related to their work on this task.
This work was supported by EPSRC grant EP/K036580/1 and a Xerox Foundation
Award.

\bibliographystyle{plainnat}
\bibliography{../a.bib,../kmh.bib}

\end{document}